# Kinematic and Dynamic Analyses of the Orthoglide 5-axis

R. Ur-Rehman, S. Caro, D. Chablat, P. Wenger
Institut de Recherche en Communication et Cybernétique de Nantes,
UMR CNRS 6597, 1 rue de la Noë, 44321 Nantes, France
Raza.Ur-Rehman@irccyn.ec-nantes.fr

*Abstract*—This paper deals with the kinematic and dynamic analyses of the Orthoglide 5-axis, a five-degree-of-freedom manipulator. It is derived from two manipulators: i) the Orthoglide 3-axis; a three dof translational manipulator and ii) the Agile eye; a parallel spherical wrist. First, the kinematic and dynamic models of the Orthoglide 5-axis are developed. The geometric and inertial parameters of the manipulator are determined by means of a CAD software. Then, the required motors performances are evaluated for some test trajectories. Finally, the motors are selected in the catalogue from the previous results.

## I. Introduction

PARALLEL kinematics machines become more and more popular in industrial applications [1, 2].This growing attention is inspired by their essential advantages over serial manipulators that have already reached the dynamic performance limits. In contrast, parallel manipulators are claimed to offer better accuracy, lower mass/inertia properties, and higher structural stiffness (i.e. stiffness-to-mass ratio) [3]. These features are induced by their specific kinematic structure, which resists to the error accumulation in kinematic chains and allows convenient actuators location close the manipulator base. Besides, the links work in parallel against the external force/torque, eliminating the cantilever-type loading and increasing the manipulator stiffness [4]. The latter makes them attractive for innovative machine-tool architectures [5, 6], but practical utilization for the potential benefits requires development of efficient kinematic and dynamic analyses, which satisfy the computational speed and accuracy requirements of relevant design procedures.

This paper focuses on the kinematic and dynamic analyses of the Orthoglide 5-axis, a spatial parallel-kinematics machine (PKM) developed for high speed operations [7]. To evaluate the forces and torques that have to be exerted by the actuators of the Orthoglide 5-axis, its kinematic and dynamic analyses are of primary importance.

## II. Orthoglide 5-axis

The Orthoglide 5-axis, illustrated in Fig. 1, is derived from a 3-dof translating manipulator, the Orthoglide 3-axis and a 2-dof spherical wrist [7].

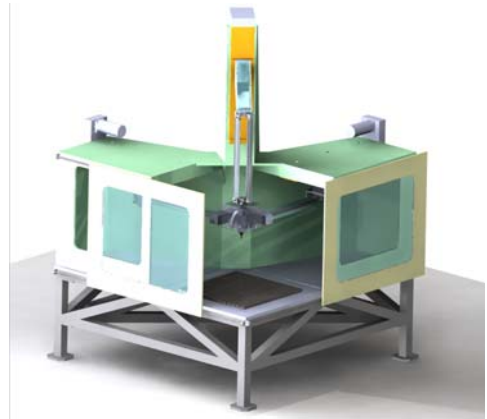

Fig. 1. The Orthoglide 5-axis

The Orthoglide 3-axis is a Delta-type PKM [8] dedicated to 3-axis rapid machining applications developed at the Research Institute in Communications and Cybernetics of Nantes (IRCCyN) [9]. This mechanism is composed of three identical legs. Each leg is made up of a prismatic joint, a revolute joint, a parallelogram joint and another revolute joint. The first joint, i.e. the prismatic joint of each leg, is actuated while the end-effector is attached to the other end of each leg. Hence, the Orthoglide 3-axis is a PKM with movable foot points and constant chain lengths.

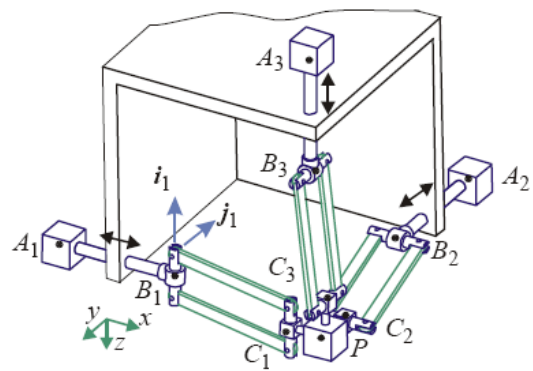

Fig. 2. The Orthoglide 3-axis

The Orthoglide 3-axis gathers the advantages of both serial and parallel kinematic architectures such as regular workspace, homogeneous performances, good dynamic performances and stiffness. The interesting features of the Orthoglide 3-axis are large regular dextrous workspace, uniform kinetostatic performances, good compactness [10] and high stiffness [11].

The two-dof spherical wrist that is implemented in the

Orthoglide 5-axis is derived from the Agile Eye, a three-dof spherical wrist developed by Gosselin and Hamel [12]. Nevertheless, the two-dof spherical wrist was designed in order to have high stiffness, [13]. A CAD model of the wrist is shown in Fig. 3. It consists of a closed kinematic chain composed of five components: the proximal 1, the proximal 2, the distal, the terminal and the base. These five links are connected by means of revolute joints, the two revolute joints connected to the base being actuated. Let us notice that the revolute joints axes intersect.

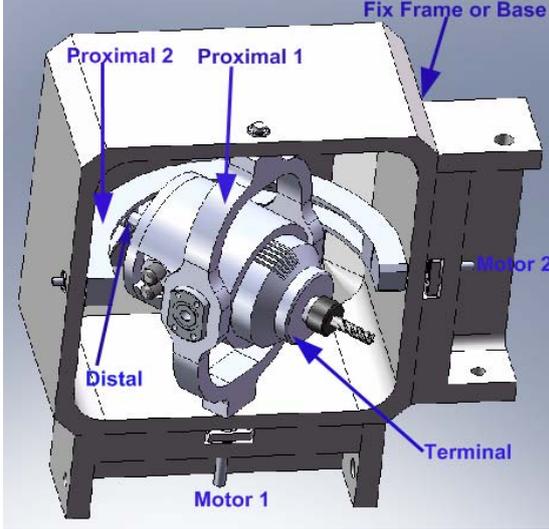

Fig. 3. Spherical wrist of the Orthoglide 5-axis

### III. TRAJECTORY PLANNING

In order to analyze the kinematic and dynamic performance of the Orthoglide 5-axis, two test trajectories are proposed. The inverse kinematic and dynamic problems of the manipulator are solved for both trajectories.

The Orthoglide 5-axis is designed for a 500x500x500 mm$^3$ cubic workspace. Let us notice that the cubic workspace center, i.e., point $I$, and the origin $O$ of the reference frame, and the intersection of the prismatic joints axes, do not coincide as explained in [9] and shown in Fig. 4. The vector $\overrightarrow{OI}$, expressed by $\mathbf{dr} = [dx\ dy\ dz]^T$, is also called position vector of the cubic workspace center. Here, the geometric centre of the path of the test trajectories is supposed to be point $I$. Accordingly, the test trajectories are defined as follows:

- Traj. *I*: semi-circular trajectory in a plane perpendicular to *XY*-plane defined by radius $R$, trajectory angle $\psi$ with *Y*-axis, trajectory plane orientation angle $\varphi$ (angle between the trajectory plane and X-axis) and vector **v** orientation angle $\delta$ (Fig. 4). Position vector **p** and wrist orientation vector **v** are given by:

$$\mathbf{p} = \begin{bmatrix} p_x \\ p_y \\ p_z \end{bmatrix} = \begin{bmatrix} dx + R\cos\psi\cos\varphi \\ dy + R\cos\psi\sin\varphi \\ dz + R\cos\psi \end{bmatrix}$$

$$\mathbf{v} = \begin{bmatrix} v_x \\ v_y \\ v_z \end{bmatrix} = \begin{bmatrix} -\cos\delta\cos\varphi \\ -\cos\delta\sin\varphi \\ -\sin\delta \end{bmatrix}$$

where $\delta$ varies from $\pi/6$ to $5\pi/6$ and $\psi$ varies from 0 to $\pi$..

- Traj. *II*: circular trajectory in horizontal or *XY*-plane defined by radius $R$, vector **v**, constant orientation angle $\gamma$ with *Z*-axis and the angle $\psi$ (Fig. 5). Position vector **p** and wrist orientation vector **v** are given by:

$$\mathbf{p} = \begin{bmatrix} p_x \\ p_y \\ p_z \end{bmatrix} = \begin{bmatrix} dx + R\cos\psi \\ dy + R\sin\psi \\ dz \end{bmatrix}$$

$$\mathbf{v} = \begin{bmatrix} v_x \\ v_y \\ v_z \end{bmatrix} = \begin{bmatrix} \sin\gamma\sin\psi \\ \sin\gamma\cos\psi \\ -\cos\gamma \end{bmatrix}$$

where $\psi$ varies from 0 to $2\pi$

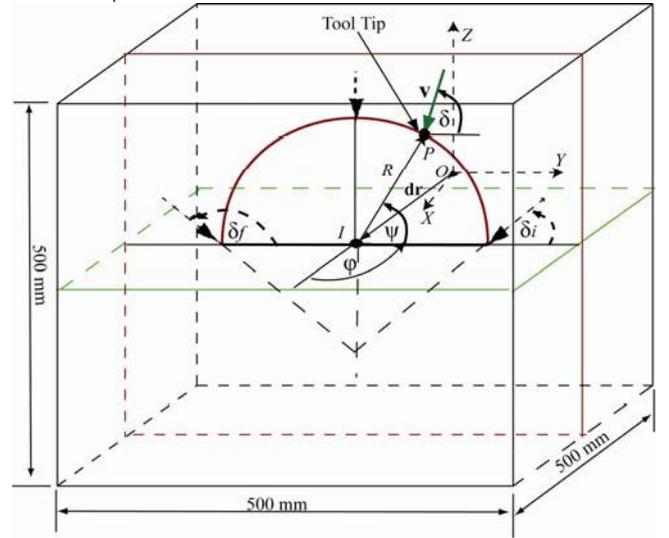

Fig. 4. Orientation of vector **v** (Traj *I*)

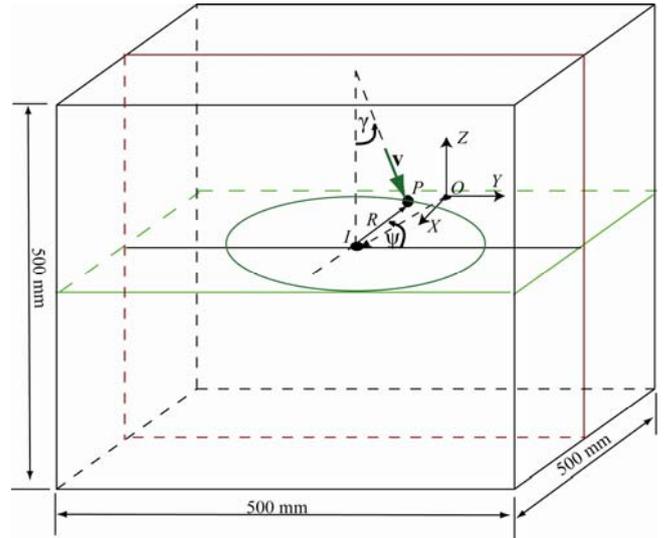

Fig. 5. Orientation of vector **v** (Traj *II*)

Let us notice that several test trajectories can be derived from Traj. *I* and Traj. *II*, by changing their parameters.

## IV. KINEMATIC ANALYSIS

### A. Orthoglide 3 axis

The geometric parameters of the Orthoglide 3-axis are defined as a function of the size of a prescribed cubic Cartesian workspace, that is free of singularities and internal collision

The kinematic architecture of the Orthoglide-3axis is shown in Fig. 2 where $A_1B_1$, $A_2B_2$ and $A_3B_3$ represent the prismatic joints and $P$ is the end-effector. Due to its Delta-linear architecture, the Orthoglide-3axis is a translating parallel manipulator with 3-DOF.

A simplified model of the Orthoglide 3-axis is illustrated in Fig. 6 [14] in which three links of length $L$ are connected by means of a spherical joint to end-effector "$P$" at one end and to the corresponding prismatic joints "$A_i$" at the other end. $\theta_x$, $\theta_y$ and $\theta_z$ are the angles between the links and the corresponding prismatic joints axes. The reference frame is coincident with the prismatic joint axes; it is the origin being the intersection point of those axes. The input position vector of the prismatic joint variables is represented by $\boldsymbol{\rho} = (\rho_x, \rho_y, \rho_z)$ and the output position vector of the end-effector by $\mathbf{p} = (p_x, p_y, p_z)$.

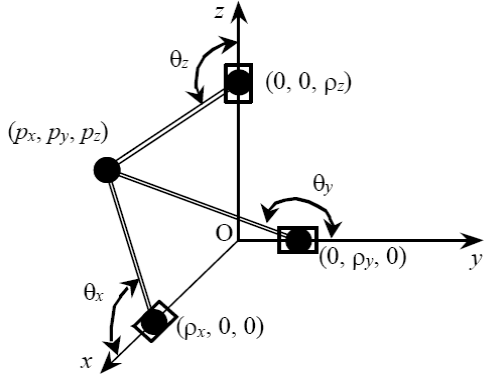

Fig. 6. Simplified model of the Orthoglide 3-axis

Using these notations, the inverse kinematic relations for a spherical singularity free workspace can be written as [10]

$$\rho_x = p_x + \sqrt{L^2 - p_y^2 - p_z^2}$$
$$\rho_y = p_y + \sqrt{L^2 - p_x^2 - p_z^2}$$
$$\rho_z = p_z + \sqrt{L^2 - p_x^2 - p_y^2}$$

Due to the Orthoglide geometry and manufacturing technology, the displacement of its prismatic joints is bounded [10], namely,

$$0 \leq \rho_{x,y,z} \leq 2L$$

The kinematic performance of the Orthoglide 3-axis is analyzed by means of the foregoing test trajectories. The velocity of end-effector $P$ throughout the trajectory, is supposed to be constant i.e. $V_p = 1$ m/s. Accordingly the actuated prismatic joints position, rates and acceleration are plotted in Figs. 7 to 9 for both test trajectories, the radius of their path being equal to 0.2 m.

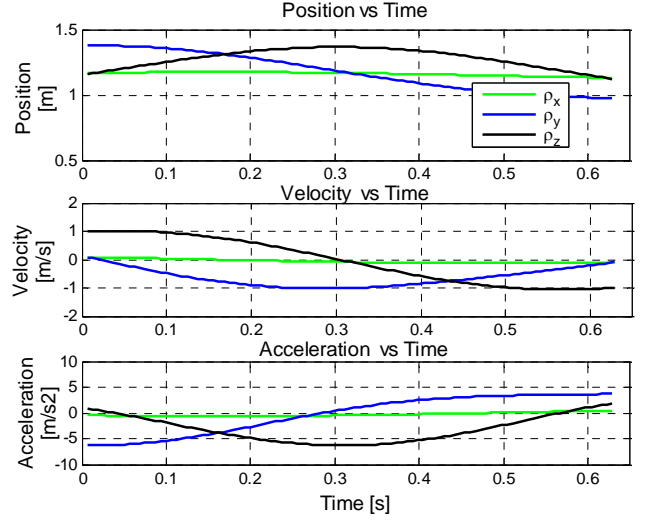

Fig. 7. Actuated prismatic joints position, rates and acceleration of the Orthoglide 3-axis for Traj *I* with $\varphi=90°$

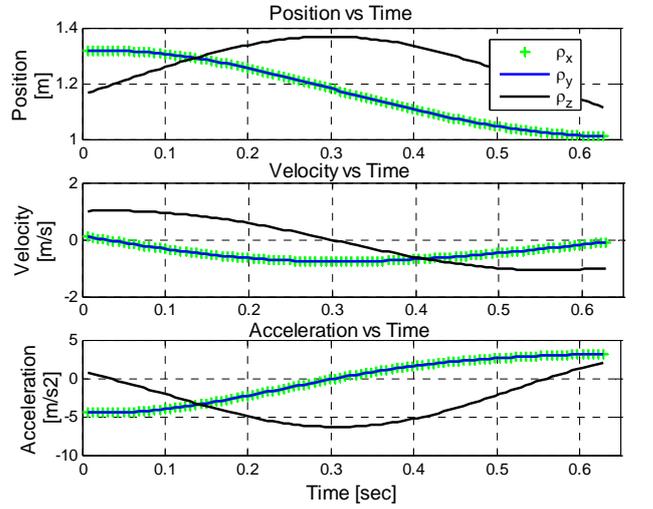

Fig. 8. Actuated prismatic joints position, rates and acceleration of the Orthoglide 3-axis for Traj *I* with $\varphi=45°$

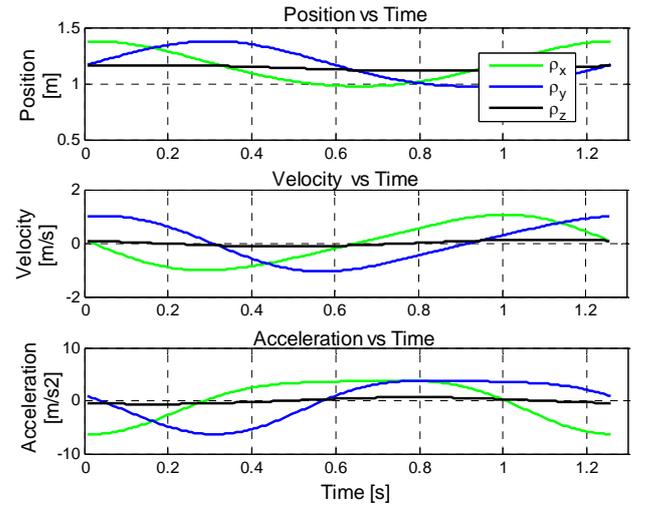

Fig. 9. Actuated prismatic joints position, rates and acceleration of the Orthoglide 3-axis for Traj *II* with $\gamma=45°$

Fig. 7 shows the kinematic performance required by the prismatic actuators when end-effector *P* moves in *YZ*-plane. Even if *P* does not move along *X*-axis, the displacement of the prismatic actuator mounted along *X*-axis is not null. Fig. 8 and Fig. 9 display the required kinematic performance of the motors when *P* follows Traj *I* ($\varphi=45°$) and Traj *II* ($\gamma=45°$), respectively for *R*=0.2m. The maximum velocities and accelerations of the prismatic actuators required for the three test trajectories are shown in Table I. We can notice that the maximum prismatic joint velocity is equal to 1 m/s whereas its maximum acceleration is equal to 6.31 m/s$^2$.

TABLE I.
MAXIMUM PRISMATIC JOINTS RATES AND ACCELERATIONS

| Test Trajectory | Max Absolute Velocity [m/s] | | | Max Absolute Acceleration [m/s$^2$] | | |
|---|---|---|---|---|---|---|
| | $V_x$ | $V_y$ | $V_z$ | $A_x$ | $A_y$ | $A_z$ |
| Traj *I*, $\gamma=90°$ | 0.12 | 1.01 | 1.06 | 0.62 | 6.32 | 6.32 |
| Traj *I*, $\gamma=45°$ | 0.77 | 0.77 | 1.07 | 4.51 | 4.51 | 6.33 |
| Traj *II*, $\gamma=45°$ | 1.06 | 1.06 | 0.12 | 6.31 | 6.31 | 0.66 |

*B. Spherical Wrist*

The spherical wrist mechanism of the Orthoglide 5-axis consists of a closed kinematic chain composed of five components: proximal-1, proximal-2, distal, terminal and the base. These five links are connected by means of revolute joints, of which axes intersect. Besides, only the two revolute joints connected to the base of the wrist are actuated. The distal has an imaginary axis of rotation passing through the intersection point of other joint axis and perpendicular to the plane of proximal-2.

The kinematic equations of the wrist are written by means of six reference frames attached to the six rigid bodies and the corresponding Denavit-Hartenberg (DH)-parameters. Fig. 10 shows the orientation of these reference frames while vector **v** represents the orientation of the terminal of the wrist, i.e. the cutting tool. Moreover, $\alpha_0$ denotes the angle between $\mathbf{e_1}$ and $\mathbf{e_2}$ while $\alpha_i$ denotes the angle between $\mathbf{e_i}$ and $\mathbf{e_{i+2}}$ ($i = 1…4$).

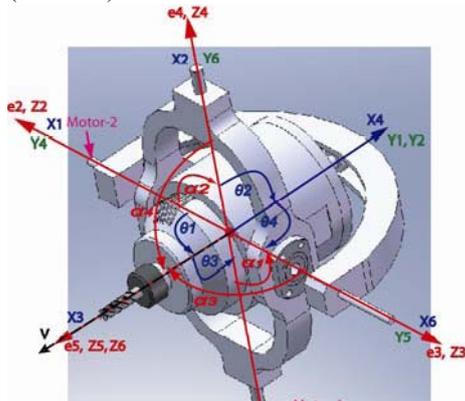

Fig. 10. Orientations of reference frames for the Orthoglide Wrist

The reference frame $R_1$ is defined in such a way that the $Z_1$-axis coincides with $\mathbf{e_1}$ and $\mathbf{e_2}$ lies in the $X_1Z_1$-plane. Similarly $R_2$ has its $Z_2$-axis in the direction of $\mathbf{e_2}$ and $\mathbf{e_1}$ lies in the $X_2Z_2$-plane. Reference frame $R_i$ ($i = 3, 4, 5, 6$) with $Z_i = \mathbf{e_i}$ are defined by the rotation of frame $R_{i-2}$ and following the DH conventions.

Finally, the inverse kinematic problem of the wrist can be derived from the definitions of the reference frames and unit vectors to develop the relations between the joints angles ($\theta_1, \theta_2, \theta_3, \theta_4$) [15]. Fig. 11 to Fig. 13 display the revolute joints angles, rates and accelerations for the two trajectories introduced in Section III, the radius of their path being equal to 0.2 m.

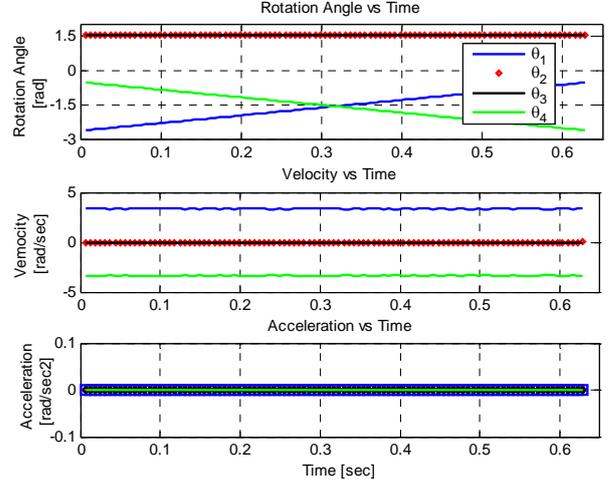

Fig. 11. Revolute joints angles, rates and accelerations of the wrist (Traj *I*, $\varphi=90°$)

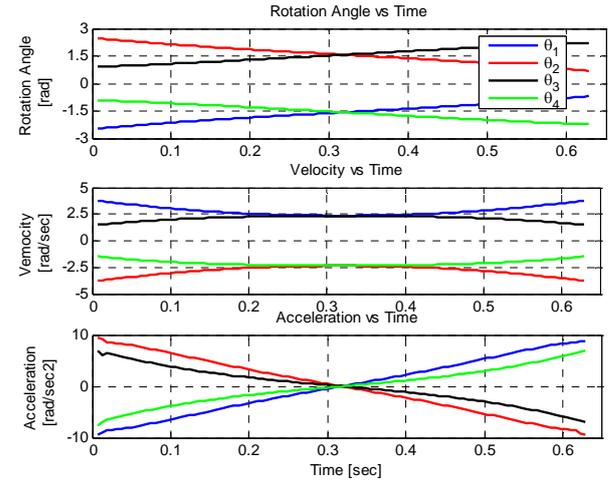

Fig. 12. Revolute joints angles, rates and accelerations of the wrist (Traj *I*, $\varphi=45°$)

Fig. 11 is the case where wrist end-effector moves in the *YZ*-plane ($\varphi=90°$) so only one of the wrist actuator ($\theta_1$) works, while Fig. 12 and Fig. 13 represent the cases where both of the actuators work. Compared to Traj *I*, both wrist actuators experience greater velocities and accelerations for Traj *II* (max velocity=5 m/s and max acceleration=22 m/s$^2$). This can be explained by the higher order variations of rotation angles in Traj *II* to that of linear variation of rotation angles in Traj *I*

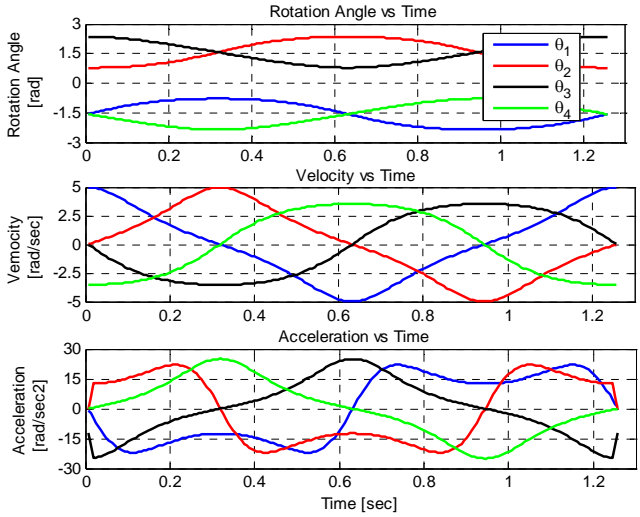

Fig. 13. Revolute joints angles, rates and accelerations of the wrist (Traj *II*, γ=45˚)

## V. Dynamic Analysis

### A. Orthoglide 3-axis

The dynamic analysis of the Orthoglide 3-axis is performed in order to evaluate the torques required by the three actuated prismatic joints. Here, we take advantage of the dynamic model developed in [16]. The geometric and dynamic parameters used in the analysis are obtained from SYMORO+ (SYmbolic MOdeling of Robots), a software for the automatic generation of symbolic model of robots [17] and defined in [16]. Fig. 14 illustrates a leg of the Orthoglide with the definition of the parameters and the frames attached to all the bodies [16].

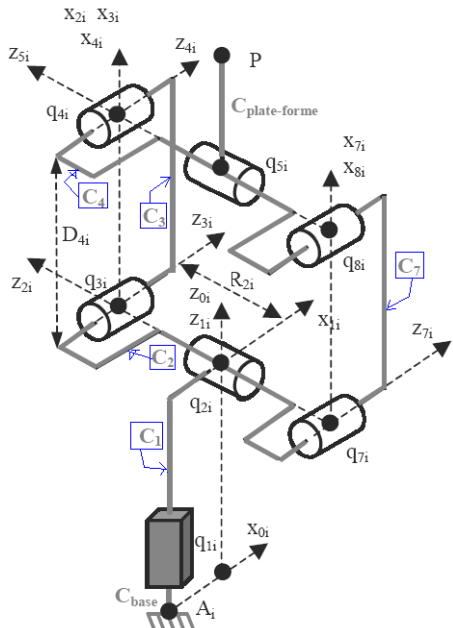

Fig. 14. Orthoglide leg parameterization for the dynamic analysis

The geometric, mass and inertial parameters used in the dynamic model were determined for the Orthoglide 5-axis by means of SolidWorks CAD software, and from the geometry of the mechanism. The dynamic performance of the Orthoglide 3-axis is then evaluated for different test trajectories. The actuators forces required to follow those trajectories are shown in Fig. 15.

TABLE II.
PARAMETERS OF ORTHOGLIDE 3-AXIS REQUIRED FOR DYNAMIC ANALYSIS

| Parameters for leg $i$ | Symbol |
|---|---|
| Mass of the platform (wrist) | $m_p$ |
| Mass of body 1 | $m_{1i}$ |
| Mass of bodies 2 and 4 | $m_{2i}, m_{4i}$ |
| Mass of bodies 3 and 7 | $m_{3i}, m_{7i}$ |
| Length of parallelograms | $d_{4i}$ |
| Width of parallelograms | $2\,r_{2i}$ |
| Actuators moment of inertia | $I_{mi}$ |

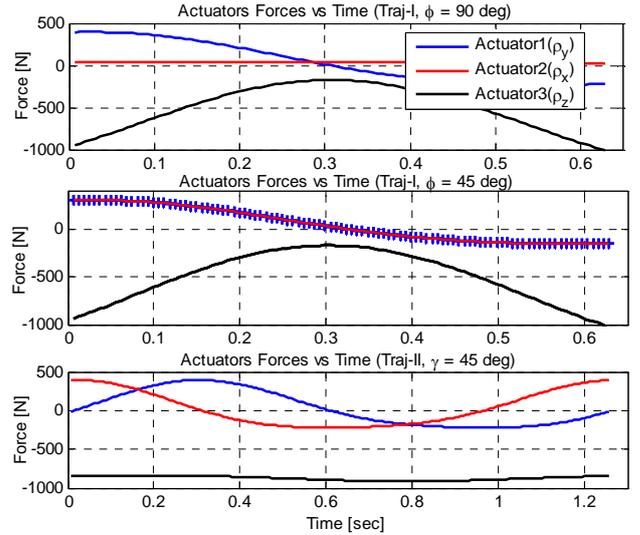

Fig. 15. Orthoglide 3-axis actuators forces (Traj-*I* and *II*)

### B. Spherical Wrist

A Newton approach is used to come up with the dynamic modeling of the Orthoglide wrist. A similar methodology was also used in [15]. Let us assume that:

- friction forces are neglected;
- there is a spherical joint is between the distal and the terminal link in order to get an isostatic mechanism;
- there is a planar joint between the distal and the proximal-2.

Thus, the free body diagrams of terminal, distal, proximal-1 and proximal-2 can be drawn. The equilibrium equations are written for each free body diagram and then, the equations used to evaluate the actuators torques are obtained [15]. The latter are shown in Fig. 16. It can be seen that for *YZ*-plane trajectory (Traj *I*, φ = 90˚), only the first actuator, aligned with the *Y*-axis (Fig. 4) experiences the torque, while for other two test trajectories both actuators work and experience torques. Second actuator, aligned with *X*-axis, experiences greater torque compared to the first actuator for Traj *I*, φ = 45˚ and for Traj *II*

In order to verify the results obtained with the Newton approach, the principle of virtual work is used.

As a matter of fact, variations in kinetic and potential energies i.e. ΔKE and ΔPE are evaluated during a time interval *dt*. The total energy variation *ΔE* over *dt* is defined as ΔE = ΔKE + ΔPE. Therefore, the total virtual work *W* is calculated by the product of the mean torques and the corresponding angular displacement during each time interval *dt*. i.e.,

$$W = \bar{T}_1.\Delta\theta_1 + \bar{T}_1.\Delta\theta_2$$

$\bar{T}_1$ and $\bar{T}_2$ being the mean torques of actuators 1 and 2, respectively during *dt, respectively*.

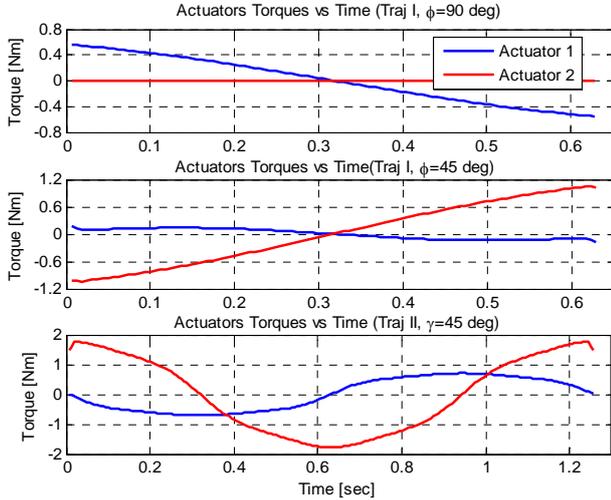

Fig. 16. Spherical wrist actuators torques (Traj-*I* and *II*)

It is noteworthy that the difference between the global virtual work *W* and *ΔE* should be null due to energy conservation. Accordingly, we compute the difference between *W* and *ΔE* and check the dynamic model of the wrist, i.e., ΔW= ΔE-W. This difference is highlighted in Fig. 17 for the three test trajectories that are considered in the scope of the study. It turns out ΔW is null in all cases. Consequently, the dynamic model of the wrist makes sense.

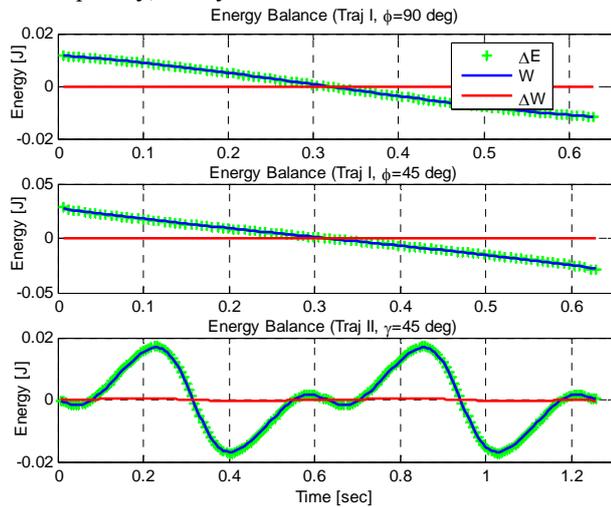

Fig. 17. Energy balance for wrist dynamics (Traj-I and II)

## VI. CONCLUSION

This paper dealt with the kinematic and dynamic analyses of the Orthoglide 5-axis, a five-degree-of-freedom manipulator. First, it turned out that kinematic and dynamic analyses of the translating part and the spherical wrist of the manipulators can be decoupled. The geometric and inertial parameters of the manipulator were determined by means of a CAD software. We came up with the dynamic model of the spherical wrist by means of a Newton approach. Besides, this model has been checked with the principle of virtual work. Then, the required motors performances were evaluated for some test trajectories. Various simulations results showed that the FFA 20-80 harmonic drive motors of 0.8 kW and the NX430 EAF motors of 1.8 kW, primarily selected for the wrist and Orthoglide 3-axis respectively, are suitable for the prototype of the Orthoglide 5-axis. In future works, friction forces as well as payload will be considered in the dynamic analysis and further test trajectories will be performed.